\documentclass[10pt,twocolumn,letterpaper]{article}

\usepackage{cvpr}              
\usepackage{amsfonts}
\usepackage{amssymb}
\usepackage{dsfont}
\usepackage{bbding}
\usepackage{algorithmic}
\usepackage{algorithm}
\usepackage{threeparttable}

\definecolor{cvprblue}{rgb}{0.21,0.49,0.74}
\usepackage[pagebackref,breaklinks,colorlinks,allcolors=cvprblue]{hyperref}

\def\paperID{11118} 
\def\confName{CVPR}
\def\confYear{2025}

\title{Accelerating Diffusion Transformer via Increment-Calibrated Caching with Channel-Aware Singular Value Decomposition} 

\author{
Zhiyuan Chen \quad Keyi Li \quad Yifan Jia \quad Le Ye \quad Yufei Ma\thanks{Corresponding Author} \\
Peking University, Beijing, China\\
{\tt\small yufei.ma@pku.edu.cn}
}

\begin{document}

\maketitle

\begin{abstract}
Diffusion transformer (DiT) models have achieved remarkable success in image generation, thanks for their exceptional generative capabilities and scalability. Nonetheless, the iterative nature of diffusion models (DMs) results in high computation complexity, posing challenges for deployment. Although existing cache-based acceleration methods try to utilize the inherent temporal similarity to skip redundant computations of DiT, the lack of correction may induce potential quality degradation. In this paper, we propose increment-calibrated caching, a training-free method for DiT acceleration, where the calibration parameters are generated from the pre-trained model itself with low-rank approximation. To deal with the possible correction failure arising from outlier activations, we introduce channel-aware Singular Value Decomposition (SVD), which further strengthens the calibration effect. Experimental results show that our method always achieve better performance than existing naive caching methods with a similar computation resource budget. When compared with 35-step DDIM, our method eliminates more than 45$\%$ computation and improves IS by 12 at the cost of less than 0.06 FID increase. Code is available at https://github.com/ccccczzy/icc. 

\end{abstract}    
\begin{figure*}[!h]
  \centering
   \includegraphics[width=1\linewidth]{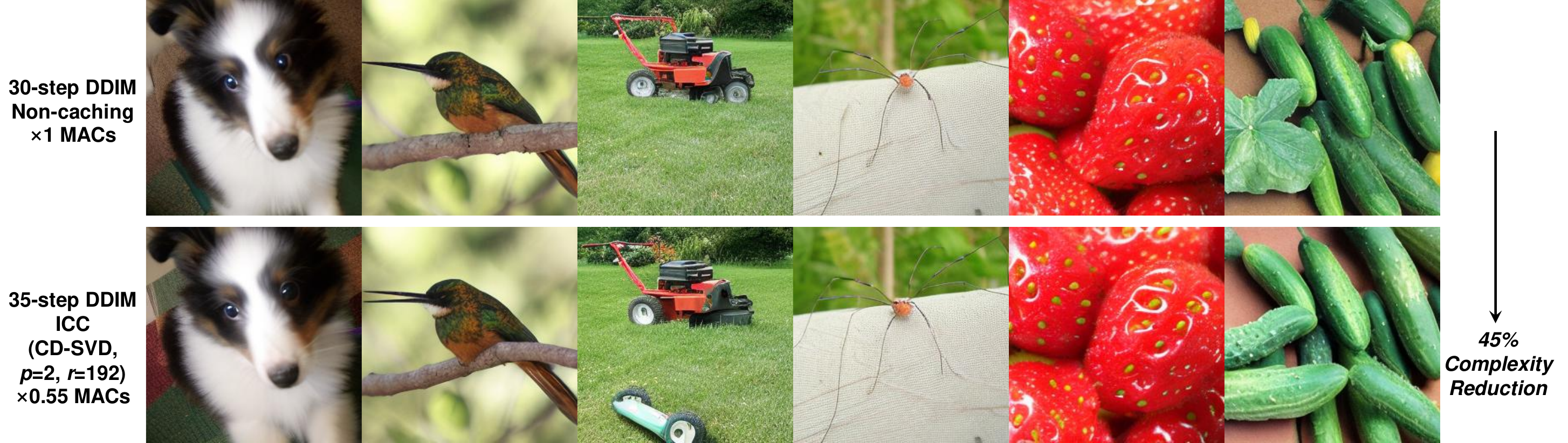}
   \caption{Visualization of increment-calibrated caching on DiT-XL/2.}
   \label{fig:vis}
\end{figure*}

\section{Introduction}
\label{sec:intro}
Diffusion models (DMs)~\cite{ncsm, ddpm, ddim, iddpm} have emerged as one of the most popular generative models, exhibiting great generation quality and diversity especially in the area of contiguous modalities, including images~\cite{ldm}, videos~\cite{mav}, and audios~\cite{wavegrad}. Throughout the reverse process of DMs, expected contents are gradually synthesized with a multi-step refinement. Although initial DMs are based on U-Net architectures~\cite{unet}, recent works have demonstrated that transformer-based DMs, \textit{i.e.}, Diffusion Transformers (DiTs) can serve as a competitive alternative~\cite{dit}. DiTs benefit from the exceptional scalability inherent in transformer architecture~\cite{transformer}, which has been extensively validated in various tasks, including image recognition~\cite{vit} and language modeling~\cite{scalinglaw}.

However, DiTs are still burdened with the iterative nature inherent in DMs. This means their high-fidelity synthesis results usually come at the cost of substantial latency, exacerbating deployment difficulties and necessitating efficiency improvement. The existing strategies to accelerate DiT models can be categorized into three primary approaches: timestep reduction, weight-oriented approximation~\cite{qdit}, and cache-based acceleration~\cite{fora, l2c}. Research works within the first approach are devoted to reducing the required reverse steps, including adopting fast samplers~\cite{dpm, dpmpp, ddim, sa} or distilling the original trajectory into fewer timesteps~\cite{cm, progressive}. 

\begin{figure}[htb]
  \centering
   \includegraphics[width=1\linewidth]{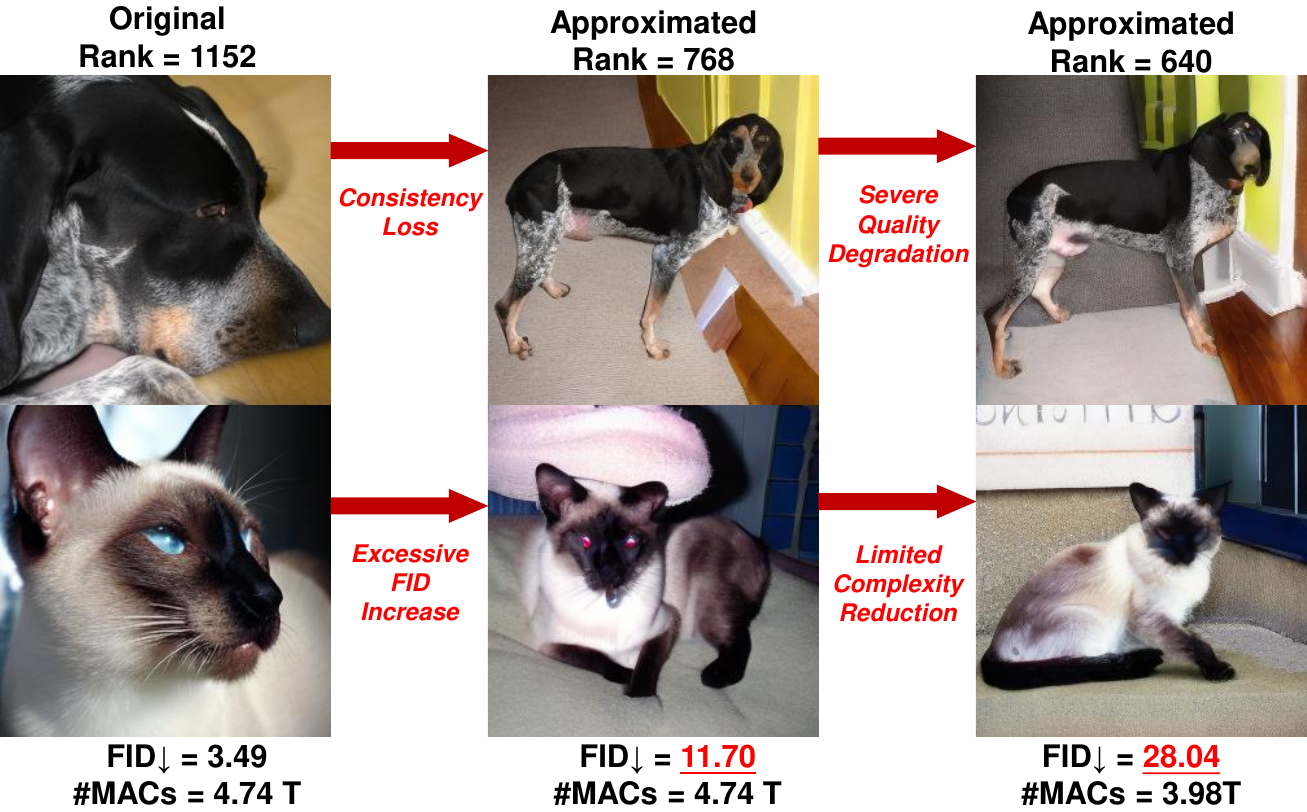}
   \caption{Degraded generation results of directly performing SVD on DiT-XL/2 at the resolution of 256 $\times$ 256.}
   \label{fig:direct_svd}
\end{figure}

Methods that fall under the category of weight-oriented approximation are inspired by the general methodology to decrease the complexity of neural networks. These works typically concentrate on eliminating redundant weight elements through pruning~\cite{diff_pruning}, adopting low-bit quantization~\cite{ptqd, qdit}, or employing low-rank approximation such as Singular Value Decomposition (SVD). However, enhancing generation efficiency by decreasing the model size of DiTs is a non-trivial task, especially when retraining is not feasible. Regarding the experimental results of SVD-based approximation, as shown in \cref{fig:direct_svd}, even when approximately 33$\%$ of the ranks are removed across all layers, the computational overhead does not diminish accordingly. Moreover, the generated images suffer from substantial quality degradation and lose the consistency with the original model.

Cache-based acceleration techniques are proposed to leverage the multi-step property of diffusion process contrarily and exploit the inherent temporal similarities to bypass redundant computations~\cite{fora, l2c, harmonica}. In this case, the intermediate features of previous timesteps are cached and directly used to approximate the outputs of subsequent timesteps without any corrections. Current related works primarily focus on designing an optimal caching mechanism, while ignoring the calibration of the cached value. This oversight may lead to quality drop and constrain the full potential of cache-based acceleration.

In this paper, we present increment-calibrated caching, a novel training-free method to accelerate DiT that distinguishes itself from aforementioned three approaches. It can been seen as a complementary combination of weight-oriented approximation and cache-based acceleration, instead of roughly packing these two techniques together. Unlike methods that directly convert the original model into a reduced version with SVD, we treat the low-rank approximated weights as calibration parameters. During the reverse process, these calibration parameters are leveraged to refine the cached values derived from the original model through an increment term, instructing them towards the desired accurate results. This ensures that increment-calibrated caching achieves a superior balance between efficiency and performance compared to conventional naive caching mechanisms.

Furthermore, we substitute the original SVD with channel-aware SVD, a refined variant that takes the channel sensitivity into consideration. In contrast to the original version, 
channel-aware SVD strengthens the calibration effect by placing greater emphasis on activation-sensitive channels. This is crucial because, as observed in various transformer-based models \cite{asvd, smoothquant}, the outlier issues in DiT can adversely affect inference results.
Main contributions of this paper are summarized as follows,
\begin{itemize}
\item We introduce increment-calibrated caching, a training-free method tailored for the acceleration of DiT. To the best of our knowledge, it is the pioneering work that aims to calibrate the cache of DMs in a training-free manner.
\item We propose channel-aware SVD, a refined SVD variant that improves increment-calibrated caching for DiT and mitigates outlier issues that could undermine the calibration effect.
\item The proposed methods are validated on state-of-the-art DiT model DiT-XL/2~\cite{dit} for conditional image synthesis and for PixArt-$\alpha$~\cite{pixartalpha} text-to-image generation. Experimental results show that our proposed method largely outperform naive caching methods with a similar computation budget. Compared with 35-step DDIM, our method eliminates more than 45$\%$ computation and improves IS by 12 at the cost of less than 0.06 FID increase.
\end{itemize}


\section{Related Works}
\noindent{\bf Diffusion Transformer.} 
While early diffusion models (DMs) relied heavily on U-Net architectures, recent works has predominantly shifted towards transformer-based DMs, specifically Diffusion Transformers (DiTs). The introduction of DiT \cite{dit}, initially applied to class-conditional image synthesis, demonstrated the scalability and effectiveness of transformer architectures within the diffusion framework. Subsequent works, such as PixArt-$\alpha$ \cite{pixartalpha} and its successors \cite{pixartdelta, pixartsigma}, have further adapted and refined the original DiT by incorporating cross-attention mechanisms and merged adaptive layer normalization, thereby extending its capabilities to text-to-image tasks. Notably, the latest iteration of Stable Diffusion \cite{sd3} has also turned to the multimodal DiT architecture. In the realm of video generation, the emergence of SORA \cite{sora} exemplifies the potential for establishing a world simulator based on DiT, further highlighting the versatility and promise of DiT.

\noindent{\bf Cache-based Acceleration.} 
Although the iterative nature of DMs leads to significant complexity, it also brings with a unique opportunity for cache-based acceleration, leveraging the temporal similarity between consecutive denoising steps. This similarity allows for the caching and reuse of intermediate results, thereby bypassing redundant computations. Some early works are devoted to exploiting the redundancy of U-Net-based DMs. For instance, DeepCache~\cite{deepcache} harnesses the temporal consistency of high-level features to streamline the computation in the main branch of U-Nets. Similarly, Block Caching \cite{cmiyc} introduces a dynamic cache mechanism that relies on the accumulated input changes.

More recent works have shifted focus towards applying these principles to DMs based on transformer architectures. FORA \cite{fora}, for example, periodically stores and retrieves the outputs of all attention and MLP layers. Meanwhile, L2C~\cite{l2c} and HarmoniCa~\cite{harmonica} aim to determine the optimal caching strategy via learning-based approaches. $\Delta$-DiT \cite{deltadit} accelerates the rear DiT blocks during the early sampling stages and the front DiT blocks in the later stages. Lastly, TokenCache \cite{tokencache} and ToCa \cite{toca} adopt a token-wise caching approach instead of layer-wise caching by distinguishing the importance of different tokens.

\noindent{\bf Low-rank Approximation.}
Low-rank approximation techniques, such as Singular Value Decomposition (SVD) and its variants, have found widespread application in the compression and acceleration of neural networks~\cite{learnrank}. Recently, researchers have explored the use of SVD in the context of Large Language Models (LLMs). For instance, FWSVD \cite{fwsvd} incorporates Fisher information to weigh the importance of parameters prior to applying SVD. ASVD \cite{asvd} addresses the outlier issues by scaling the weight matrix based on activation distribution. Additionally, SVD-LLM \cite{svdllm} establishes a direct correlation between singular values and inference loss. However, there has been limited exploration of SVD's application to the acceleration of DMs. Our experimental results, presented in \cref{fig:direct_svd}, demonstrate that a direct application of low-rank approximation to DiT significantly degrades the generation quality.

\section{Methodology}

\begin{figure*}[htb]
  \centering
   \includegraphics[width=1\linewidth]{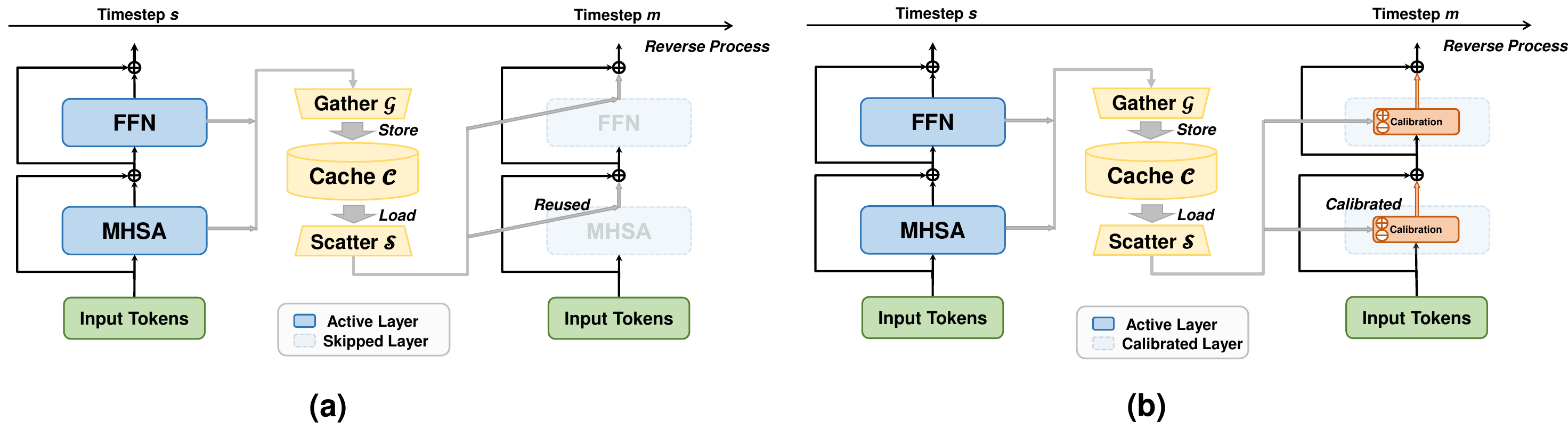}
   \caption{Overview of (a) naive caching and (b) proposed increment-calibrated caching. The former stores intermediate results of previous denoising steps and directly reuse in later timesteps which may induce unavoidable error. The proposed increment-calibrated caching corrects the cached value with calibration parameters approximated from model itself.}
   \label{fig:overview}
\end{figure*}

\subsection{Preliminary}
The basic framework of DMs consists of two parts, \textit{i.e.}, the forward process and reverse process. The former illustrates the procedure that gradually turns clean data $z_0$ into gaussian noise $z_T\sim\mathcal{N}(0, \mathrm{}I)$ in $T$ timesteps. At each timestep $t \in [1, T]$, random noise $\epsilon_t$ is added to the current data $z_{t-1}$ with the noise scheduler factor $\beta_t$,
\begin{equation}
z_t = \sqrt{1-\beta_t}z_{t-1} + \sqrt{\beta_t}\epsilon_t
\label{eq:forward_process}
\end{equation}
Reverse process starts from $z_T$ sampling from standard normal distribution. To generate new samples obeying the real distribution, the reverse process iteratively estimates the accumulated noise and remove it from current data with a sampler $\Psi(\cdot)$. Taking DDPM~\cite{ddpm} as an example, it can be represented as,
\begin{equation}
z_{t-1} = \Psi(\epsilon_{\theta}(z_t, t))) = \frac{1}{\sqrt{\alpha_t}}(z_t-\frac{1-\alpha_t}{\sqrt{1-\overline{\alpha}_t}}\epsilon_{\theta}(z_t, t))+\sigma_t\epsilon
\label{eq:reverse_process}
\end{equation}
where $\alpha_t$, $\overline{\alpha}_t$, and $\sigma_t$ are constants depending on timestep $t$, and $\epsilon\sim\mathcal{N}(0, \mathrm{}I)$. $\epsilon_{\theta}(z_t, t)$ denotes a noise estimation network parameterized by $\theta$, which usually adopts a U-Net~\cite{ddpm, ddim, iddpm, ldm} or transformer~\cite{dit, sit, pixartalpha, hunyuandit, sd3, opensora} architecture. 

\subsection{Naive Caching for Diffusion Transformer}
DiT models are composed of stacked blocks, and each contains a Multi-Head Self-Attention (MHSA) layer and a Feed-Forward Network (FFN) layer. Then, it can be represented as $F_l(\cdot)^{L}_{l=1}$, where $L$ denotes the network depth, and $F_l$ can be either MHSA or FFN layer. Naive caching methods are proposed to reuse the results of both layer types and accelerate the reverse process, since the outputs from consecutive timesteps always exhibit inherent similarity. 

As shown in \cref{fig:overview}(a), key modules of naive caching mechanism are the cache $\mathcal{C}$, gather $\mathcal{G}$, and scatter $\mathcal{S}$. During generation process, the cache carries the intermediate outputs that may be reused in the future, which is jointly maintained by gather and scatter. Both gather and scatter can be represented as binary matrices, \textit{i.e.}, $\mathcal{G}, \mathcal{S} \in {\left \{0, 1  \right \}^{T\times L}}$, where element $\mathcal{G}_{t,l}$ and $\mathcal{S}_{t,l}$ denote whether stores or loads the results of layer $l$ at timestep $t$, respectively.

Given input $x_{s,l}$ at timestep $s$, if corresponding bit $\mathcal{G}_{s,l}$ of gather matrix is activated, layer $l$ will be fully computed to update the cache: $\mathcal{C}_l(y) \leftarrow  F_l(x_{s,l})$. Then, for a subsequent timestep $m$, 
only when $\mathcal{S}_{m,l}$ does not equal 1, layer $l$ is required to be re-computed. Otherwise, it will be approximated with the cached value to bypass the computations: $y_{m,l} \approx \mathcal{C}_l(y)$. To minimize the total computation while ensuring the generation quality, the gather and scatter matrices should be carefully designed through handcrafting~\cite{fora} or learning-based approaches~\cite{l2c, harmonica}.

\subsection{Increment-Calibrated Caching with SVD}
However, the naive caching strategy usually entails the risk of quality degradation. A potential remedy to this issue is to calibrate the cached value with an increment term, thereby guiding it towards the accurate result. We initiate the discussion from a simple linear layer with weight $W \in \mathbb{R}^{C_o \times C_i}$, where $C_i$ and $C_o$ represent input and output channels, respectively. This method can be extended to MHSA or FFN by calibrating all the linear operations inside. Consistent with the notation established in the preceding subsection, it is assumed that the input $x_{s,l}$ of previous step $s$ is also synchronously stored alongside output $y_{s,l}$ as $C_l(x) \leftarrow x_{s,l}$. Consequently, the accurate result $y_{m,l}$ at timestep $m$ can be reformulated as the sum of cached output and an increment term,
\begin{equation}
y_{m,l} = \mathcal{C}_l(y) + W(x_{m,l} - \mathcal{C}_l(x)) 
\label{eq:icc}
\end{equation}

Directly computing the increment term leads to unchanged $\mathcal{O}(NC_{i}C_{o})$ complexity compared with full computation, where $N$ is the token number. To obtain a trade-off between efficiency and performance, low-rank approximation is conducted on $W$ based on SVD. With SVD, the weight matrix $W$ can be factorized into three matrices: $U$, $\Sigma$, and  $V^T$, such that $W=U \Sigma V^T$. The diagonal matrix $\Sigma$ 
consists of positive singular values sorted in the descending order. The truncation approximation keeps the largest $r$ singular values to minimize the reconstruction error with a rank-$r$ constraint. Then, $U$, $\Sigma$, $V^T$ are replaced with $U_r \in \mathbb{R}^{C_o \times r}$, $\Sigma_r \in \mathbb{R}^{r \times r}$, $V^T_r \in \mathbb{R}^{r \times C_i}$, approximating $W$ as $W_r = U_r \Sigma_r V_r^T$ = $W^a_rW^b_r$. $W^a_r \in \mathbb{R}^{C_o \times r}$ and $W^b_r \in \mathbb{R}^{r\times C_i}$ are kept as the calibration parameters. Thus, the complexity is reduced to $\mathcal{O}(N(C_i+C_o)r)$, and the cached output can be corrected as,
\begin{equation}
y_{m,l} \approx \mathcal{C}_l(y) + {W^a_r}{W^b_r}(x_{m,l} - \mathcal{C}_l(x))
\label{eq:icc_svd}
\end{equation}

\subsection{Improved Calibration with Channel-Aware SVD}
As demonstrated in prior works \cite{fwsvd, asvd, svdllm}, the straightforward application of SVD-based low-rank approximation may induce unacceptable performance degradation of neural networks, particularly when only a limited number of ranks are retained. 
Original SVD methods solely focus on the static weights and fail to take the influence of input activations into consideration.
Thus, this results in a mismatch between the reconstruction error and the actual model performance. 
\begin{figure}[htb]
  \centering
   \includegraphics[width=1\linewidth]{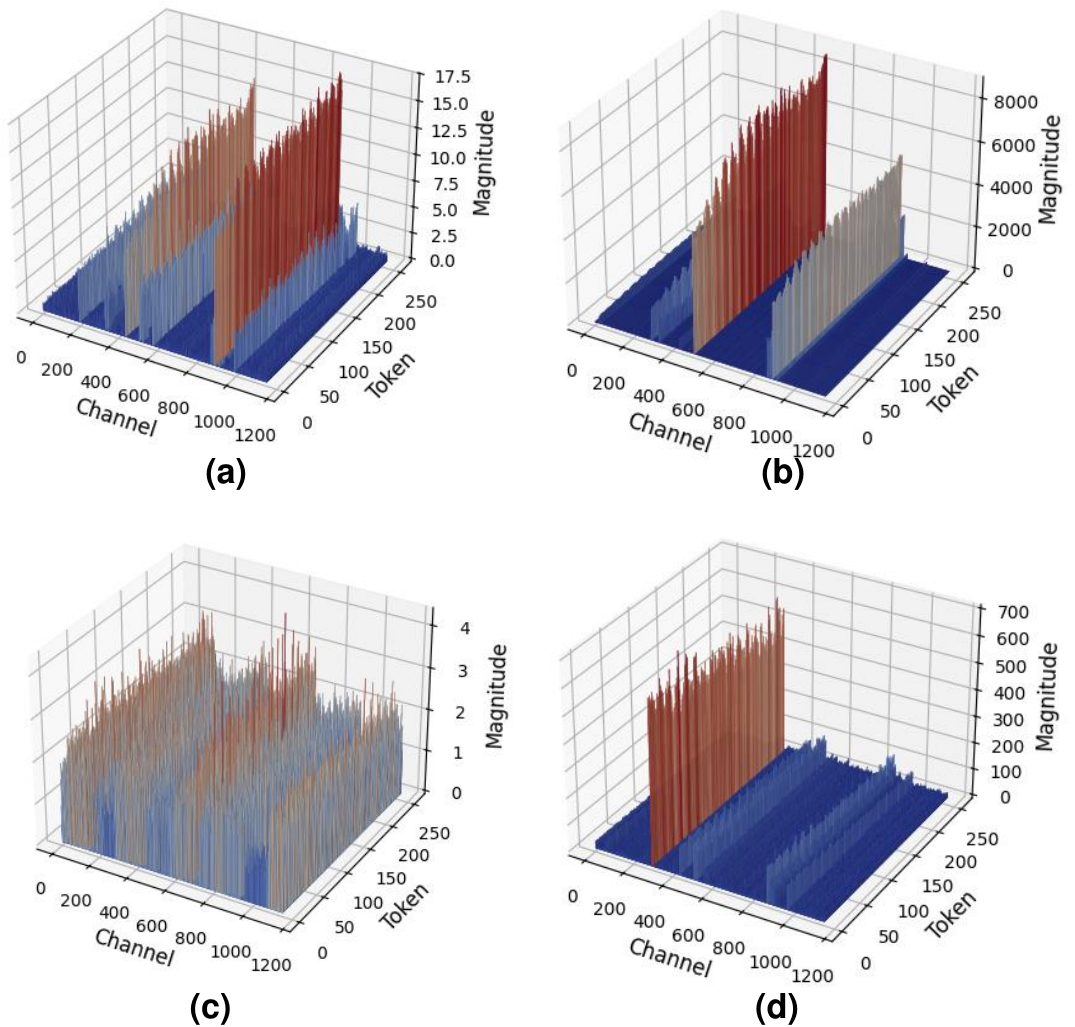}
   \caption{Outlier issues of DiT models appearing in (a) input channel of FFN FC1, (b) output channel of FFN FC2, (c) input channel, and (d) output channel of MHSA output projection.}
   \label{fig:outlier}
\end{figure}
For increment-calibrated caching of DiT, this issue may become even more severe. \cref{fig:outlier} illustrates the activation distribution across all the linear layers within a single block of DiT-XL/2~\cite{dit}. Notably, outliers are evident in both the input and output channel dimensions. This implies that even minor reconstruction errors can significantly weaken the effectiveness of calibration in certain channels. Therefore, during the correction process, it is imperative to give heightened attention to the weight elements of these sensitive channels.

To this end, we adopt channel-aware SVD, which takes the sensitivity of both input and output channels into consideration when approximating a given weight matrix. Assuming the importance of different channels can be weighed by two diagonal matrices $S_i \in \mathbb{R}^{C_i \times C_i}$ and $S_o \in \mathbb{R}^{C_o \times C_o}$, where each element corresponds to one input or output channel, respectively. Then, the weight matrix $W$ can be scaled to $W'$ to take the channel sensitivity into consideration,
\begin{equation}
W = S_o^{-1}(S_oWS_i)S^{-1}_i=S_o^{-1}W'S^{-1}_i
\label{eq:icc_svd}
\end{equation}
To reduce its complexity, $W'$ is factorized with SVD and approximated into $U'_r{\Sigma'}_r{V'}_r^T$ by only keeping the first $r$ singular values. Then, we re-scale the approximated $W'$ with the inverse of matrix $S_i$ and $S_o$ to obtain the approximation of original matrix $W$. It can be represent as the multiplication of matrix ${\hat{W}}^a_r \in \mathbb{R}^{C_o \times r}$ and ${\hat{W}}^b_r \in \mathbb{R}^{r \times C_o}$, which will be used as calibration parameters,    
\begin{equation}
W \approx (S_o^{-1}U'_r){\Sigma'}_r({V'}_r^TS^{-1}_i)={{\hat{W}}^a_r}{{\hat{W}}^b_r}
\label{eq:rescale}
\end{equation}
To obtain the matrix $S_o^{-1}$ and $S_i^{-1}$, we propose two feasible methods: channel-activation-aware SVD (CA-SVD) and channel-delta-aware SVD (CD-SVD). The former is a extended version of ASVD~\cite{asvd}, where the mean magnitude of activations across channels is calculated based on a small calibration dataset and used as the scale. CD-SVD utilizes the average inter-timestep difference to generate the scale matrices, as shown in \cref{alg:delta_svd}. 

\begin{algorithm}[!h]
    \caption{Channel-delta-aware Singular Value Decomposition}
    \label{alg:delta_svd}
    \renewcommand{\algorithmicrequire}{\textbf{Input:}}
    \renewcommand{\algorithmicensure}{\textbf{Output:}}
    
    \begin{algorithmic}[1]
        \REQUIRE Calibration set $\mathcal{D}$, DiT model $\epsilon_{\theta}(\cdot)$, number of timesteps $T$, sampler $\Psi(\cdot)$, noise scheduler $\beta_t$
        \ENSURE Scale matrix $S_i$ and $S_o$ of each linear layer $l$ of $\epsilon_{\theta}(\cdot)$
        \FOR{each linear layer $l$ of $\epsilon_{\theta}(\cdot)$}
            \STATE Initialize zero vector $S_i$ and $S_o$     
        \ENDFOR
        
        \FOR{each $z_0 \in \mathcal{D}$}
            \STATE $t \sim \mathcal{U}[2, T]$
            \STATE $z_t \sim \mathcal{N}(z_t, \alpha_tz_0, \sigma_t^2)$ 
            \STATE $z_{t-1} \leftarrow \Psi(\epsilon_{\theta}(z_t, t))$ and
                \FOR{each linear layer $l$ of $\epsilon_{\theta}(\cdot)$}
                \STATE Cache both inputs and outputs
                \ENDFOR
            \STATE $z_{t-2} \leftarrow \Psi(\epsilon_{\theta}(z_{t-1}, t-1))$ and
                \FOR{each linear layer $l$ of $\epsilon_{\theta}(\cdot)$}
                \STATE Compute input and output differences with cache, accumulate them to $S_i$ and $S_o$
                \ENDFOR
        \ENDFOR
        \STATE Turn $S_i$ and $S_o$ into diagonal matrices
        \RETURN Outputs
    \end{algorithmic}
\end{algorithm}
\section{Experiment}
\subsection{Experimental Setup}
\noindent{\bf Models, Datasets, and Metrics.} To demonstrate the effectiveness of the proposed method, evaluation is conducted based on commonly used Diffusion Transformer model, DiT-XL/2~\cite{dit} for class-conditional image synthesis and PixArt-$\alpha$~\cite{pixartalpha} for text-to-image generation. The default samplers are DDIM~\cite{ddim} and DPM~\cite{dpm}, respecitively. 50,000 images from 1,000 classes in ImageNet~\cite{imagenet} are sampled at the resolution of 256$\times$256, evaluating performance with Inception Score (IS)~\cite{is}, Fréchet Inception Distance (FID)~\cite{fid}, sFID, Precision~\cite{prec}, and Recall. 
For text-to-image generation, we employ 30,000 captions randomly sampled from MSCOCO-2014~\cite{coco} to generate images. We use FID-30k and Clip Score~\cite{clipscore} to assess the image quality and text-image alignment, respectively. 
The computation complexity is measured with the number of Multiply-and-Accumulate (MAC) operations, which is the basic unit of computation intensive matrix multiplications.
                      
\noindent{\bf Caching Pattern Setting.} 
The naive caching pattern of FORA~\cite{fora} is selected as the baseline, where the outputs of all MHSA and FFN layers are synchronously cached according to a fixed period $p$. Thus, the gather matrix $\mathcal{G}$ is set as, $\mathcal{G}_{t,l}= \mathds{1}_{p|t}$, $\forall t \in [1, T]$, $l \in [1, L]$. The scatter matrix $\mathcal{S}$ can be seen as the complement of $\mathcal{G}$: $\mathcal{S}_{t,l} = 1 - \mathcal{G}_{t,l}$. Besides, it should be noticed that, the proposed calibration method can not only be applied to this particular case, but is compatible with other naive caching mechanism of the format defined in subsection 3.2. 

\noindent{\bf Calibration Mechanism Setting.} 
To simplify discussion, we employ a unified rank $r$ for all linear layers to generate the calibration parameters. To conduct CA-SVD or CD-SVD, 256 images are randomly selected from the training set as the calibration set.

\subsection{Comparison with Naive Caching}
\begin{table*}[htb]
  \centering
  \begin{tabular}{c|cc|ccccc|c}
    \toprule
    Method  &Sampler &\#Steps     
    &IS$\uparrow$   &FID$\downarrow$    &sFID$\downarrow$   &Prec.$\uparrow$  &Recall$\uparrow$ 
    &\#MACs$\downarrow$(T)\\ 
    \midrule
    DiT-XL/2 (cfg = 1.5)  &DDPM &250
    &278.2 &2.27 &4.60 &0.83 &0.57   
    &59.31 \\
    \midrule
    DiT-XL/2 (cfg = 1.5)  &DDIM &40  
    &239.9 &2.35 &4.26 &0.80 &0.59   
    &9.49\\
    Naive Caching ($p$ = 2) &DDIM &80        
    &243.4 &2.29 &4.46 &0.81 &0.59 
    &9.51\\
    Naive Caching ($p$ = 3) &DDIM &120        
    &242.7 &2.33 &4.62 &0.81 &0.59      
    &9.53\\
    \textbf{ICC (CA-SVD, $r$ = 128, $p$ = 2)} &DDIM &66        
    &244.0 &2.23 &4.35 &0.80 &0.59           
    &9.40\\
    \textbf{ICC (CD-SVD, $r$ = 192, $p$ = 2)} &DDIM &62        
    &251.7 &\textbf{2.10} &\textbf{4.25} &0.81 &\textbf{0.59}           
    &9.40\\
    \textbf{ICC (SVD, $r$ = 256, $p$ = 2)} &DDIM &58        
    &\textbf{257.8} &2.14 &4.29 &\textbf{0.82} &0.58           
    &\textbf{9.35}\\
    \midrule
    DiT-XL/2 (cfg = 1.5) &DDIM &30  
    &235.0 &2.64 &4.38 &0.80 &0.59   
    &7.12\\
    Naive Caching ($p$ = 2) &DDIM &60        
    &240.0 &2.48 &4.60 &0.81 &0.59 
    &7.13\\
    Naive Caching ($p$ = 3) &DDIM &90        
    &243.8 &2.52 &4.91 &0.81 &0.57      
    &7.15 \\
    \textbf{ICC (CA-SVD, $r$ = 128, $p$ = 2)} &DDIM &50        
    &241.7 &2.31 &4.33 &0.80 &\textbf{0.60}           
    &7.11\\
    \textbf{ICC (CD-SVD, $r$ = 192, $p$ = 2)} &DDIM &46        
    &253.0 &\textbf{2.28} &\textbf{4.26} &0.82 &0.57           
    &\textbf{6.98}\\
    \textbf{ICC (SVD, $r$ = 256, $p$ = 2)} &DDIM &44     
    &\textbf{258.7} &2.30 &4.31 &\textbf{0.83} &0.57           
    &7.09\\
    \midrule
    DiT-XL/2 (cfg = 1.5)  &DDIM &20  
    &224.3 &3.49 &4.93 &0.79 &0.58   
    &4.75\\
    Naive Caching ($p$ = 2) &DDIM &40        
    &235.4 &2.95 &4.93 &0.80 &0.57 
    &4.75\\
    Naive Caching ($p$ = 3) &DDIM &60        
    &234.4 &3.00 &5.54 &0.80 &0.57      
    &4.76 \\
    \textbf{ICC (CA-SVD, $r$ = 128, $p$ = 2)} &DDIM &32        
    &236.8 &2.77 &4.35 &0.80 &\textbf{0.58}           
    &4.55\\
    \textbf{ICC (CD-SVD, $r$ = 192, $p$ = 2)}  &DDIM &30        
    &250.6 &\textbf{2.53} &\textbf{4.34} &0.81 &0.57           
    &4.55\\
    \textbf{ICC (SVD, $r$ = 256, $p$ = 2)} &DDIM &28        
    &\textbf{257.1} &2.66 &4.47 &\textbf{0.82} &0.56           
    &\textbf{4.51}\\
    \midrule
    DiT-XL/2 (cfg = 1.5) &DDIM &10  
    &160.4 &12.33 &11.34 &0.67 &0.51    
    &2.37\\
    Naive Caching ($p$ = 2) &DDIM &20        
    &193.33 &6.73 &8.77 &0.74 &0.53   
    &2.38\\
    Naive Caching ($p$ = 3) &DDIM &30        
    &196.9 &6.27 &9.48 &0.76 &0.51     
    &2.38 \\
    \textbf{ICC (CA-SVD, $r$ = 128, $p$ = 2)}  &DDIM &16        
    &205.9 &5.56 &\textbf{5.40} &0.75 &\textbf{0.55}           
    &2.27\\
    \textbf{ICC (CD-SVD, $r$ = 192, $p$ = 2)} &DDIM &14       
    &217.6 &\textbf{5.05} &5.46 &0.77 &0.53           
    &\textbf{2.12}\\
    \textbf{ICC (SVD, $r$ = 256, $p$ = 2)} &DDIM &14        
    &\textbf{219.5} &5.10 &5.59 &\textbf{0.78} &0.52           
    &2.26\\
    \bottomrule
  \end{tabular}
  \caption{Comparison of naive caching and increment-calibrated caching based on DiT-XL/2 evaluated on ImageNet.}
  \label{tab:dit256}
\end{table*}

\begin{table*}[htb]
  \centering
  \begin{tabular}{c|cc|cc|c}
    \toprule
    Method  &Sampler &\#Steps     
    &FID-30k$\downarrow$    &Clip Score$\uparrow$
    &\#MACs$\downarrow$(T)\\ 
    \midrule
    PixArt-$\alpha$ (cfg = 4.5)  &DPM &30        
    &9.54 &30.66  
    &8.33\\
    \midrule
    PixArt-$\alpha$ (cfg = 4.5)  &DPM &15        
    &39.57 &29.11  
    &4.17\\
    Naive Caching ($p$ = 2) &DPM &30        
    &10.39 &30.54 
    &4.17\\
    \textbf{ICC (SVD, $r$ = 64, $p$ = 2)} &DPM &26        
    &9.80 &30.60            
    &3.94\\
    \textbf{ICC (CA-SVD, $r$ = 64, $p$ = 2)} &DPM &26        
    &\textbf{9.29} &\textbf{30.64}            
    &\textbf{3.94}\\
    \midrule
    PixArt-$\alpha$ (cfg = 4.5)  &DPM &10        
    &86.54 &27.21   
    &2.78\\
    Naive Caching ($p$ = 2) &DPM &20
    &25.47 &29.68 
    &2.78\\
    \textbf{ICC (SVD, $r$ = 64, $p$ = 2)} &DPM &18        
    &18.69 &30.04            
    &2.73\\
    \textbf{ICC (CA-SVD, $r$ = 64, $p$ = 2)} &DPM &18        
    &\textbf{13.94} &\textbf{30.29}            
    &\textbf{2.73}\\
    \bottomrule
  \end{tabular}
  \caption{Comparison of naive caching and increment-calibrated caching based on PixArt-$\alpha$ evaluated on MSCOCO-2014.}
  \label{tab:pixart}
\end{table*}

\begin{table*}[tb]
  \centering
  \begin{threeparttable}
  \begin{tabular}{c|cc|ccccc|c}
    \toprule
    Method &Sampler &\#Steps     
    &IS$\uparrow$   &FID$\downarrow$    &sFID$\downarrow$   &Prec.$\uparrow$  &Recall$\uparrow$ 
    &\#MACs$\downarrow$(T)\\ 
    \midrule
    DiT-XL/2 (cfg = 1.5)  &DDIM &40
    &239.9 &2.35 &4.26 &0.80 &0.59   
    &9.49 \\
    \midrule
    Naive Caching (L2C~\cite{l2c}) &DDIM &50
    &244.1 &\textbf{2.27} &\textbf{4.23} &0.81 &\textbf{0.59}
    &9.04\\
    \textbf{ICC (CD-SVD, $r$ = 192, $p$ = 2)} &DDIM &46        
    &\textbf{253.0} &2.28 &4.26 &\textbf{0.82} &0.57           
    &\textbf{6.98}\\
    
  \bottomrule
  \end{tabular}
  \caption{Comparison between increment-calibrated caching and L2C~\cite{l2c}.}
  \label{tab:comp_l2c}
\end{threeparttable}
\end{table*}

\noindent{\bf Unified Caching Pattern.} 
The proposed increment-calibrated caching is compared to naive caching with the same caching pattern, \textit{i.e.}, FORA~\cite{fora}. To ensure fairness, we constrain the computational overhead of all methods to be the same level and investigate their differences in performance with varying sampling settings. As shown in \cref{tab:dit256}, for class-conditional synthesis, our increment-calibrated caching achieves better results than naive caching in all performance metrics when both adopt the same FORA caching pattern, particularly under constrained computational resource budget. For instance, with a MAC limit of approximately 2.37T, the proposed increment-calibrated caching method achieves an IS score that is 22 points higher than naive caching, while also reducing FID and sFID by 1.22 and 3.37, respectively.
Similar results can also be observed in text-to-imge generation, as shown in \cref{tab:pixart}. Even without channel-aware SVD for improvement, the increment-calibrated caching outperforms naive caching in both FID and Clip Score. The application of CA-SVD further improves the results, offering 0.51 $\sim$ 4.75 FID reduction. 

\noindent{\bf Distinct Caching Pattern.}
In this work, we have not focused extensively on the design of caching patterns and have instead adopted the coarse-grained method employed by FORA~\cite{fora}.
L2C, on the other hand, is a caching-only acceleration technique that conducts fine-grained optimization, thereby further maximizing the potential of caching.
However, as depicted in \cref{tab:comp_l2c}, our proposed caching method achieves comparable performance with L2C~\cite{l2c} and reduces the computation overhead by 23$\%$, thanks to the correction effect of increment-calibration. It is also worth noting that L2C employs a learning-based method to optimize its caching pattern, necessitating additional training. 
Conversely, our method's calibration parameters are derived in a training-free manner and can be obtained in just a few minutes on a single NVIDIA 4090D GPU, significantly reducing the complexity and efforts of deployment.

\subsection{Comparison with Fast Sampling}
\begin{figure*}[bt]
  \centering
   \includegraphics[width=1\linewidth]{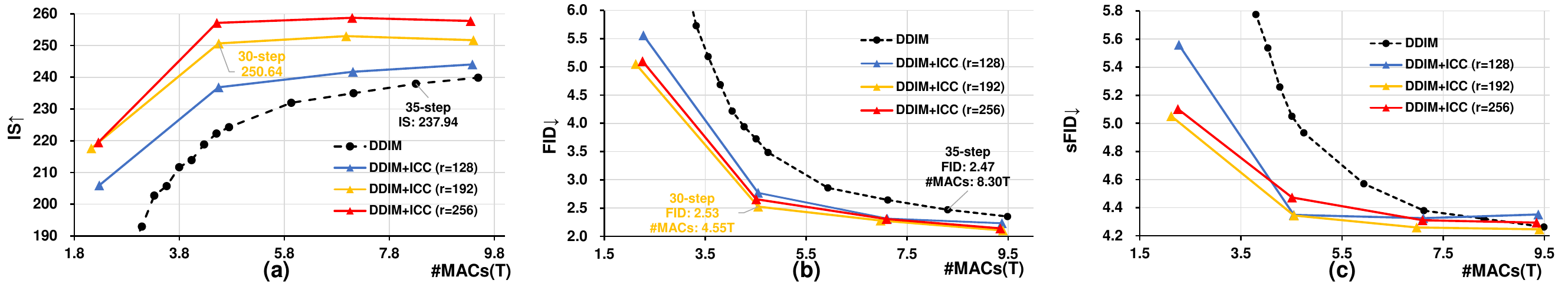}
   \caption{The trade-off between the number of MACs and (a) IS, (b) FID, and (c) sFID for class-conditional synthesis on ImageNet with DiT-XL/2. For increment-calibrated caching, the period $p$ is set to 2.}
   \label{fig:comp_ddim}
\end{figure*}

\begin{figure}[tb]
  \centering
   \includegraphics[width=1.0\linewidth]{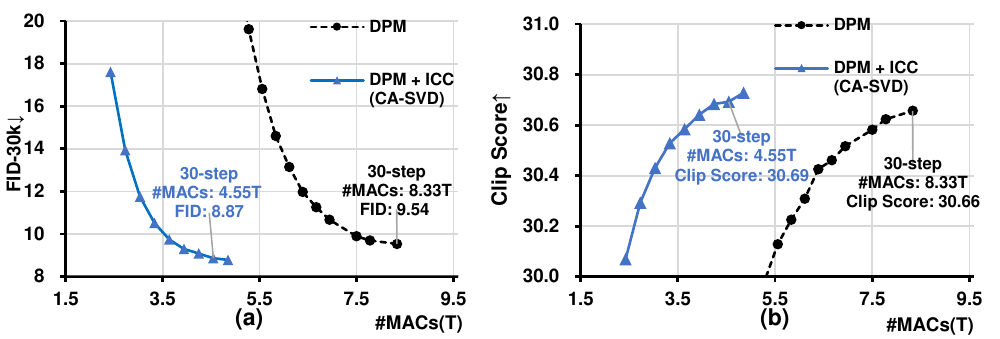}
   \caption{The trade-off of between the number of MACs and (a) FID-30k and (b) Clip Score for text-to-image generation on MSCOCO-2014 with PixArt-$\alpha$. For increment-calibrated caching, the period $p$ is set to 2 and the rank $r$ is 64.}
   \label{fig:comp_dpm_final}
\end{figure}

For DMs, a common practice to reduce computation overhead is decreasing the sampling steps at the cost of quality loss. \cref{fig:comp_ddim} and \cref{fig:comp_dpm_final} illustrates the complexity-quality trade-off curves of increment-calibrated caching and fast samplers. 
For class-conditional synthesis, our proposed method outperforms DDIM in IS, FID, and sFID when a fixed computation budget is given, especially when the complexity is constrained. That means, with increment-calibrated caching, it is hopeful to accelerate the sampling greatly depending on the lower bound of specified metric. For example, when the maximum of FID is set to around 2.5, increment-calibrated caching with 30 steps outperform 35-step DDIM with 12 IS increase at the cost of less than 0.06 FID increase. Meanwhile, the computation is reduced by 45$\%$, which means the theoretical acceleration ratio of increment-calibrated caching of relative to DDIM reaches 1.8.
For text-to-image generation, the proposed method can be used to reduce the computation overhead while maintain or even improve FID and Clip Score. For example, DPM sampler begins to converge when the number of steps reaches 30, where increment-calibrated caching can eliminate 45$\%$ MAC operations and decreases FID by 0.67. 

\subsection{Ablation Study}

\begin{table}[tb]
  \centering
  \begin{tabular}{c|ccccc}
    \toprule
    Calibration Method     
    &IS$\uparrow$   &FID$\downarrow$    &sFID$\downarrow$\\ 
    \midrule
    SVD ($r=128$) 
    &207.1 &5.11 &5.69\\
    \midrule
    CD-SVD$_i$ ($r=128$) 
    &216.9 &4.31 &5.19\\
    CD-SVD$_o$ ($r=128$)
    &209.1 &4.97 &5.51\\
    CD-SVD ($r=128$)
    &\textbf{217.3} &\textbf{4.27} &\textbf{5.13}\\
  \bottomrule
  \end{tabular}
  \caption{The comparison of CD-SVD-based increment-calibrated caching with two reduced versions: CD-SVD$_i$ and CD-SVD$_o$. Results are based on DiT-XL/2, evaluated on ImageNet wit 20-step DDIM.}
  \label{tab:comp_iochann}
\end{table}

\noindent{\bf Sensitivity of Input and Output Channels.}
This work applies scaling for both the input and output channels of weight matrices during the process of channel-aware SVD. To evaluate the sensitivity disparities between these two dimensions, we compare CD-SVD with two reduced variants: CD-SVD$_i$, which only scales the input channels, and CD-SVD$_o$, which only scales the output channels. As illustrated in \cref{tab:comp_iochann}, CD-SVD outperforms both CD-SVD$_i$ and CD-SVD$_o$ across all metrics. Notably, CD-SVD$_i$ demonstrates a significantly better performance compared to CD-SVD$_o$. This suggests that, while outlier activations are present in both input and output channels, the weight elements associated with input channels are more sensitive to this issue.

\begin{table}[tb]
  \centering
  \begin{tabular}{c|c|cccc}
    \toprule
    Calibration Method &$\#$Steps     
    &IS$\uparrow$   &FID$\downarrow$    &sFID$\downarrow$\\ 
    \midrule
    SVD ($r=128$) &32
    &227.0 &3.13 &4.71\\
    CA-SVD ($r=128$) &32
    &\textbf{236.8} &\textbf{2.77} &\textbf{4.35}\\
    CD-SVD ($r=128$) &32
    &234.1 &2.86 &4.51\\
    \midrule
    SVD ($r=192$) &30
    &244.5 &2.58 &4.34\\
    CA-SVD ($r=192$) &30
    &\textbf{252.8} &2.74 &4.69\\
    CD-SVD ($r=192$) &30
    &250.6 &\textbf{2.53} &4.34\\
    \midrule
    SVD ($r=256$) &28
    &257.1 &\textbf{2.66} &\textbf{4.47}\\
    CA-SVD ($r=256$) &28
    &261.1 &3.19 &6.48\\
    CD-SVD ($r=256$) &28
    &\textbf{261.4} &2.69 &4.57\\
  \bottomrule
  \end{tabular}
  \caption{Comparison of different calibration mechanisms with varying rank. Results are based on DiT-XL/2, evaluated on ImageNet with DDIM.}
  \label{tab:comp_cali}
\end{table}

\noindent{\bf Comparison of Calibration Mechanisms.}
In this work, the calibration of increment-calibrated caching can be achieved through three methodologies: the original SVD, CA-SVD, and CD-SVD. Our findings indicate that no single method consistently outperforms the others across all scenarios. \Cref{tab:comp_cali} presents the results for all three approaches with varying rank values. When only a limited number of ranks are retained as calibration parameters, CA-SVD and CD-SVD demonstrate superior performance compared to the original SVD. This advantage stems from their consideration of channel sensitivity, allowing them to filter the most significant ranks, which subsequently enhances the FID, sFID, and IS metrics. However, as the number of ranks used to generate calibration parameters increases, the application of either CA-SVD or CD-SVD may negatively impact both the FID and sFID, although they continue to improve the IS metric.

\begin{figure}[tb]
  \centering
   \includegraphics[width=1.0\linewidth]{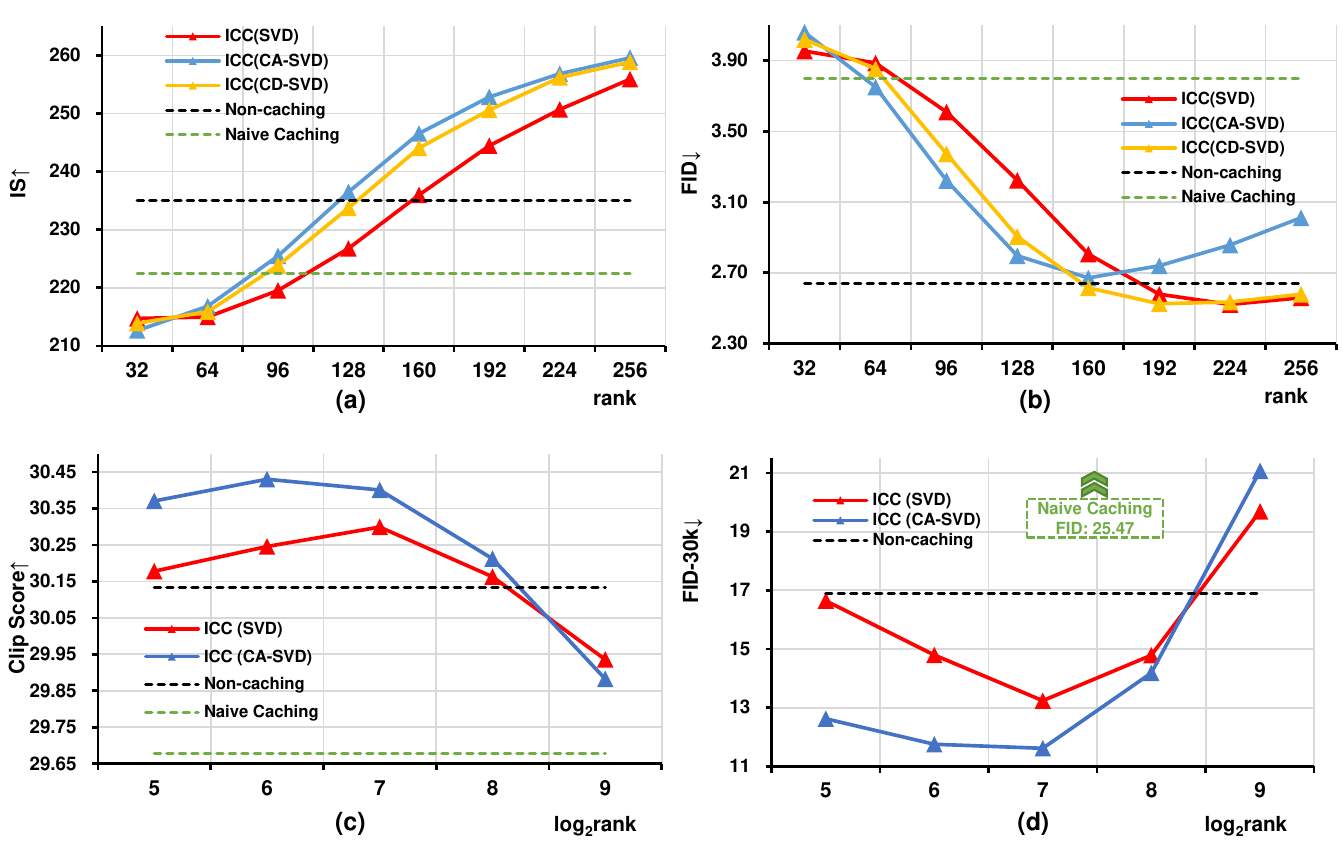}
   \caption{The impact of increased rank $r$ on (a) IS, (b) FID for DiT-XL/2 on ImageNet with 30-step DDIM, and (c) Clip Score, (d) FID-30k for PixArt-$\alpha$ on MSCOCO-2014 with 20-step DPM.}
   \label{fig:rank_inc_final}
\end{figure}

\noindent{\bf Impact of Varying Rank.}
According to simple intuition, a natural assumption is that the image quality will be improved smoothly when the rank $r$ of calibration parameter is increased. However, our experiments indicate that this is not always the truth, as shown in \cref{fig:rank_inc_final}. For results of text-to-image generation, although FID-30k decreases with the increasing $r$ at first, it begins to increase when $r$ reaches 256. Although the adoption of channel-aware SVD improves the results, but its curve is still not monotonically decreasing. 
Another related phenomenon is that some calibration settings even outperform the non-caching sampling with the same number of step. It means not all the singular values have a positive influence on the effect of increment-calibration. Thus, the proposed method has the potential to obtain significant complexity reduction without sacrificing generation quality.

\section{Conclusion}

In this work, we propose increment-calibrated caching, a novel method to reduce the computation complexity of DiT in a training-free manner. Combing the ideology of low-rank approximation and caching-based acceleration, increment-calibrated caching corrects cached activations from previous timesteps and guides them towards required accurate results in the subsequent timesteps. This process is ensured with calibration parameters that are generated from the pre-trained model itself with SVD-based low-rank approximation. We further introduce channel-aware SVD, which takes the channel sensitivity into consideration and prevents harmful outlier issues. Experimental results show that our method largely outperforms naive caching method in performance. Besides, compared with 35-step DDIM, our method eliminates more than 45$\%$ computations and improves IS by 12 at the cost of less than 0.06 FID increase.

\section*{Acknowledgements}
This work was supported in part by the National Key Research and Development Program of China under Grant 2023YFB4404603; and in part by the National Natural Science Foundation of China under Grant 92164301, Grant 62204003, Grant 62225401, and Grant 61927901.

\clearpage

{
    \small
    \bibliographystyle{ieeenat_fullname}
    \bibliography{main}
}

\clearpage
\setcounter{page}{1}
\maketitlesupplementary

\section{Detailed Illustration of CA-SVD}
\begin{algorithm}[!h]
    \caption{Channel-activation-aware Singular Value Decomposition}
    \label{alg:asvd}
    \renewcommand{\algorithmicrequire}{\textbf{Input:}}
    \renewcommand{\algorithmicensure}{\textbf{Output:}}
    \begin{algorithmic}[1]
        \REQUIRE Calibration set $\mathcal{D}$, DiT model $\epsilon_{\theta}(\cdot)$, number of timesteps $T$, noise scheduler $\beta_t$
        \ENSURE Scale matrix $S_i$ and $S_o$ of each linear layer $l$ of $\epsilon_{\theta}(\cdot)$
        \FOR{each linear layer $l$ of $\epsilon_{\theta}(\cdot)$}
            \STATE Initialize zero vector $S_i$ and $S_o$     
        \ENDFOR
        
        \FOR{each $z_0 \in \mathcal{D}$}
            \STATE $t \sim \mathcal{U}[1, T]$
            \STATE $z_t \sim \mathcal{N}(z_t, \alpha_tz_0, \sigma_t^2)$ 
            \STATE Compute $\epsilon_{\theta}(z_t, t))$ and
                \FOR{each linear layer $l$ of $\epsilon_{\theta}(\cdot)$}
                \STATE Accumulate the magnitude of inputs and outputs to $S_i$ and $S_o$ 
                \ENDFOR
        \ENDFOR
        \STATE Turn $S_i$ and $S_o$ into diagonal matrices
        \RETURN Outputs
    \end{algorithmic}
\end{algorithm}

As show in \cref{alg:asvd}, to enhance increment-calibrated caching with a extended version of ASVD~\cite{asvd}, we obtain the distribution of activations of reverse process from a calibration set, which is composed of a small number of clean images. Just like the training process of DDPM~\cite{ddpm}, the reparameterization trick is utilized to model the distribution of noised data at any given timestep, which is streamed to DiT model. During the execution of DiT model, the accumulated activation magnitude is calculated, which will used to scale weight matrix before SVD as \cref{eq:icc_svd}.

\section{Implementation Details}
We implement our methods based on the source codes and pre-trained models of DiT~\cite{dit} and PixArt-$\alpha$~\cite{pixartalpha}. All the comparative experiments are conducted under the same conditions to ensure fairness, where the hardware platform and random seed are both unified. 
For the class-conditional image synthesis on ImageNet dataset, all the results are based on 8$\times$ NVIDIA RTX 4090D GPUs with a global batch size of 128 and data precision of TF32.
For the results of text-to-image generation, we employ one 4090D GPU and set the batch size to 2. All the images are generated with the precision of FP16.

\section{Theoretical analysis}
 To explain why the proposed method works, we analyze the induced error of both naive caching and increment-calibrated caching. To simplify the discussion, we ignore the error accumulation effect. Assuming the output of linear layer $l$ at step $s$ will be reused at step $m$, the error $\Delta y$ of increment-calibrated caching can be formulated as,
\begin{equation}
    \begin{aligned}
    \Delta y ={}& Wx_{m, l}-(Wx_{s, l} + W_r(x_{m, l}-x_{s, l})) \\
    =& (W-W_r)(x_{m, l}-x_{s, l}) = \Delta W \Delta x
    \end{aligned}
\label{eq:error_icc}
\end{equation}
where $W$ is the original weight, $W_r$ is the approximated weight of rank $r$, and $x_{t, l}$ denotes the input at step $t$. The upper bound of MSE can be represented as $ \left\|\Delta W\right\|_2^2\left\|\Delta x\right\|_2^2$. According to the property of SVD, a larger $r$ always tends to decrease $\left\|\Delta W\right\|_2^2$. Note that the proposed method will be reduced to naive caching when $r$ equals $0$, therefore the proposed method can outperform naive caching by limiting the upper bound of MSE.

\begin{figure*}
\centering
   \includegraphics[width=1.0\linewidth]{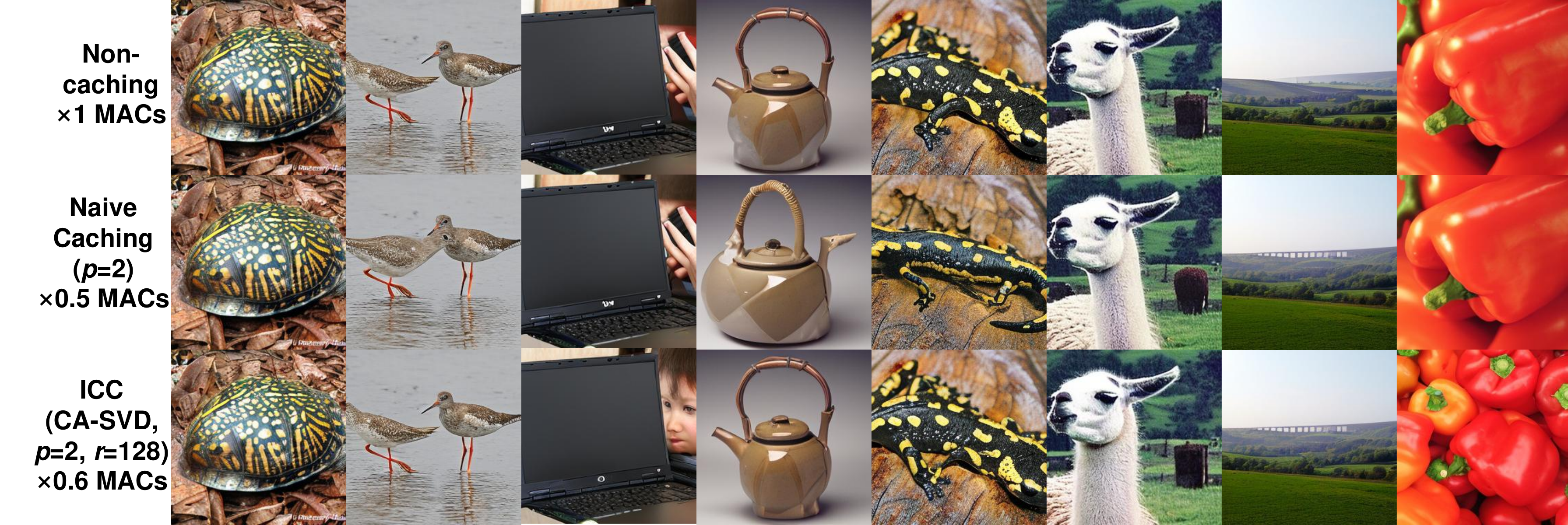}
   \caption{Visualization of the proposed method evaluated on DiT-XL/2 with 100-step DDIM. The cfg scale is set to 4.}
   \label{fig:vis_100_4}
\end{figure*}

\begin{figure*}
\centering
   \includegraphics[width=1.0\linewidth]{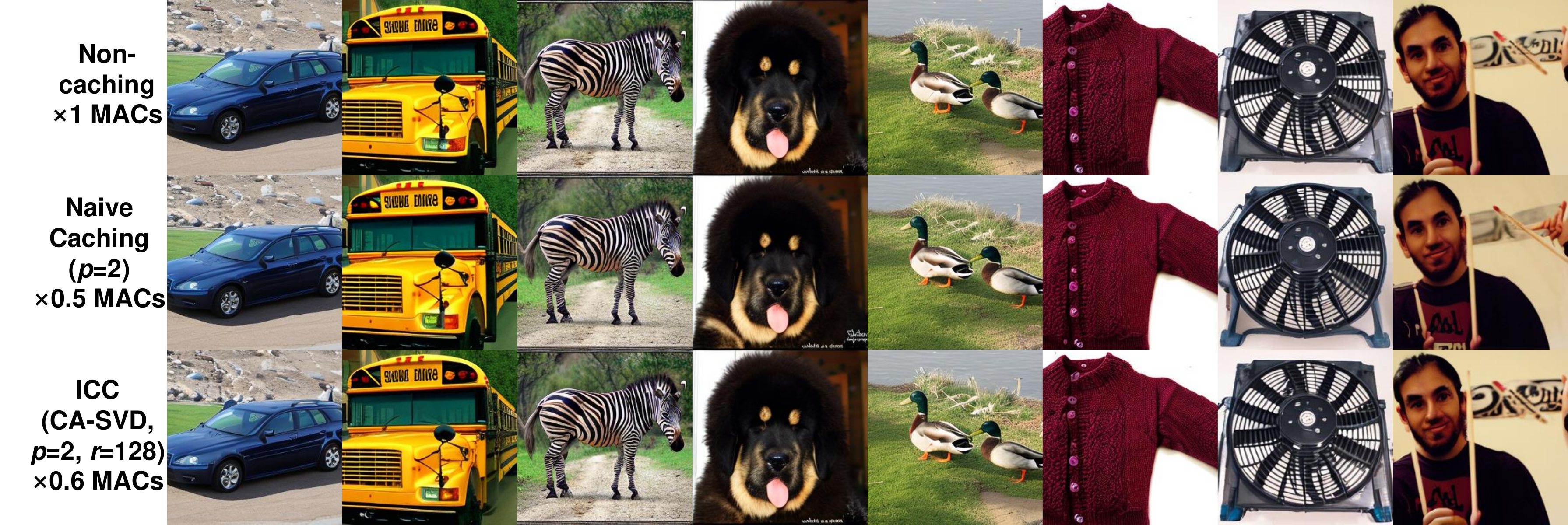}
   \caption{Visualization of the proposed method evaluated on DiT-XL/2 with 50-step DDIM. The cfg scale is set to 4.}
   \label{fig:vis_50_4}
\end{figure*}

\begin{figure*}
\centering
   \includegraphics[width=1.0\linewidth]{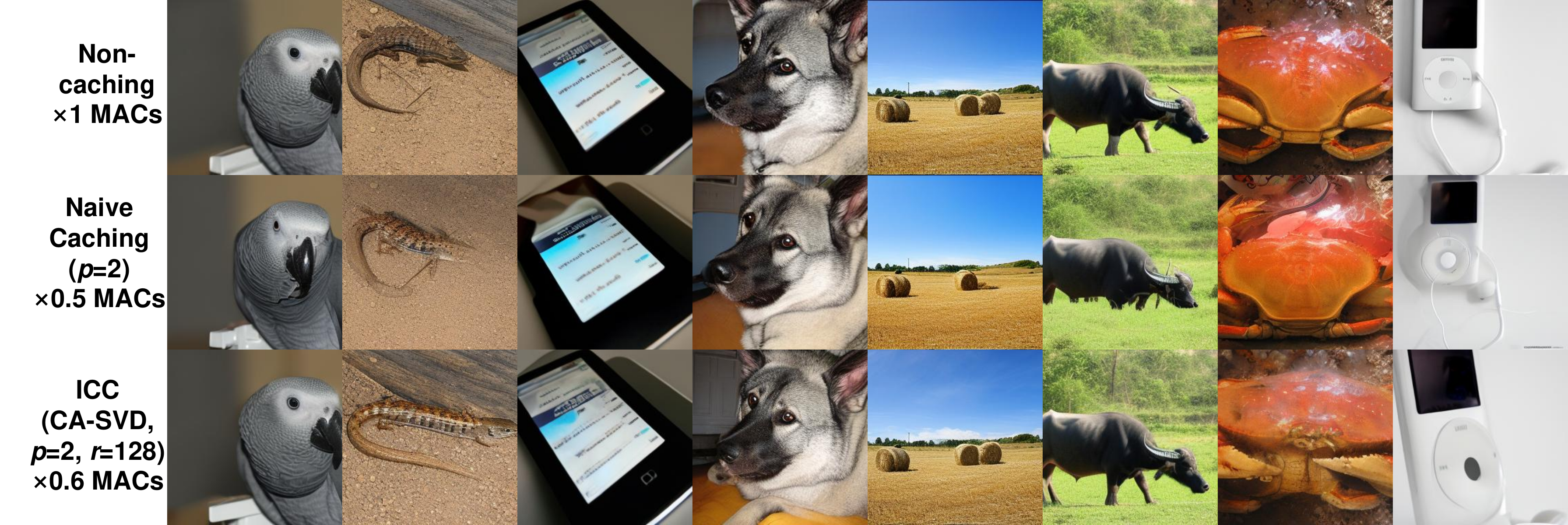}
   \caption{Visualization of the proposed method evaluated on DiT-XL/2 with 20-step DDIM. The cfg scale is set to 4.}
   \label{fig:vis_20_4}
\end{figure*}

\section{Visualization}
To visually demonstrate the effectiveness of the proposed method, We provided generation images based on non-caching, naive caching and the proposed increment-calibrated caching with different sampler settings in \cref{fig:vis_100_4}, \cref{fig:vis_50_4} and \cref{fig:vis_20_4}. We found naive caching tends to blur the details or Change the posture of an object, especially when a small step number is given. The proposed method can effectively correct these distortions and generate images of higher quality with marginal computation cost.

\end{document}